\pdfoutput=1

\documentclass[11pt]{article}

\usepackage{acl}

\usepackage{times}
\usepackage{latexsym}
\usepackage{url}
\usepackage{makecell}
\usepackage{wrapfig}
\usepackage{graphics}
\usepackage{booktabs}
\usepackage{subcaption}
\usepackage{multirow}
\usepackage{bbm}
\usepackage{amsmath,amsfonts,bm}
\usepackage{makecell}
\usepackage{array}
\usepackage[para]{threeparttable}
\usepackage{booktabs}
\usepackage{tabularx}
\usepackage{xspace}

\usepackage{tablefootnote}
\usepackage{adjustbox}
\usepackage[most]{tcolorbox}
\usepackage{colortbl}
\usepackage{pbox}
\usepackage{arydshln}

\usepackage[T1]{fontenc}

\usepackage[utf8]{inputenc}

\usepackage{microtype}
\usepackage{hyperref}

\newcommand{\dashrule}[1][black]{%
  \color{#1}\rule[\dimexpr.5ex-.2pt]{4pt}{.4pt}\xleaders\hbox{\rule{4pt}{0pt}\rule[\dimexpr.5ex-.2pt]{4pt}{.4pt}}\hfill\kern0pt%
}

\title{\textsc{Multi-Task Inference}:\\ Can Large Language Models Follow Multiple Instructions at Once?}

\author{Guijin Son$^{1,2,3}$ \quad Sangwon Baek$^{2}$ \quad Sangdae Nam$^{2,5}$ \quad Ilgyun Jeong$^{2,4}$  \quad \textbf{Seungone Kim}$^{6,7}$\thanks{Corresponding Author} \\ 
\\
Yonsei University$^{1}$ \quad EleutherAI$^{2}$ \quad OneLineAI$^{3}$ \quad Korea University$^{4}$ \\ \quad VESSL AI$^{5}$ \quad KAIST AI$^{6}$ \quad Carnegie Mellon University $^{7}$ \\
\texttt{spthsrbwls123@yonsei.ac.kr} \quad \texttt{seungone@kaist.ac.kr}} 

\begin{document}
\maketitle
\begin{abstract}

Large language models (LLMs) are typically prompted to follow a \textit{single} instruction per inference call. In this work, we analyze whether LLMs also hold the capability to handle \textit{multiple} instructions simultaneously, denoted as \textsc{Multi-Task Inference}. For this purpose, we introduce the \textsc{MTI Bench} (\textbf{M}ulti-\textbf{T}ask \textbf{I}nference \textbf{Bench}mark), a comprehensive evaluation benchmark encompassing 5,000 instances across 25 tasks. Each task in the \textsc{MTI Bench} involves 2 to 3 sub-tasks. As expected, we first demonstrate that \textsc{Multi-Task Inference} reduces the total inference time by $\times1.46$ times in average since it does not require multiple inference calls. Interestingly, contrary to the expectation that LLMs would perform better when tasks are divided, we find that state-of-the-art LLMs, such as \textsc{Llama-2-Chat-70B} and \textsc{GPT-4}, show up to 7.3\% and 12.4\% improved performance with \textsc{Multi-Task Inference} compared to \textsc{Single-Task Inference} on the \textsc{MTI Bench}. We release the \textsc{MTI Bench} dataset and our code at this link \footnote{\url{https://github.com/guijinSON/MTI-Bench}}.

\end{abstract}

\begin{figure}[t!]
\includegraphics[width=\columnwidth]{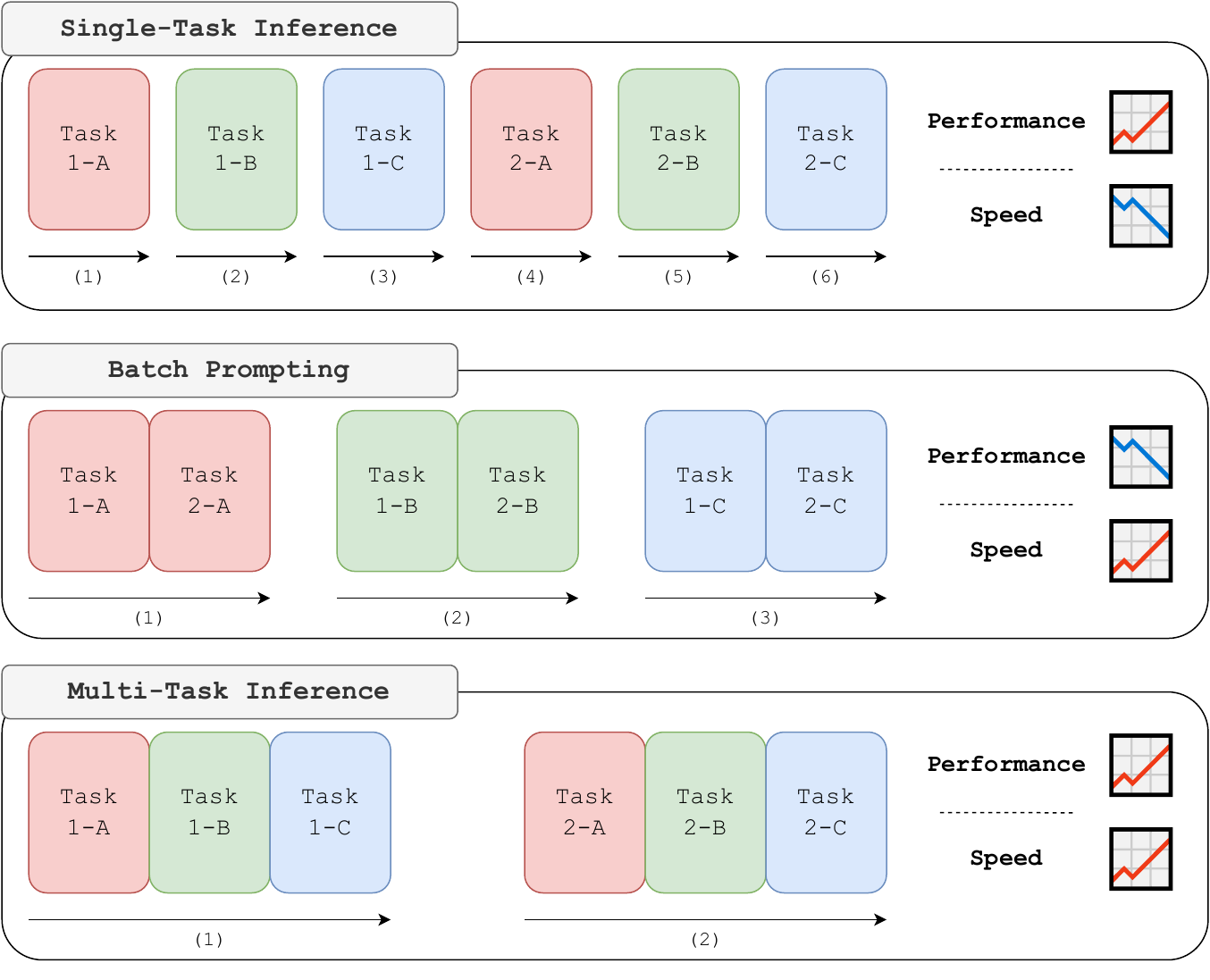}
\centering
\caption{\footnotesize Comparison of the three inference methods for handling tasks composed of three sub-tasks: \textsc{Single-Task Inference}, \textsc{Batch Prompting}, and \textsc{Multi-Task Inference}. \textsc{Multi-Task Inference} shows reliable performance as \textsc{Single-Task Inference} and provides faster speed as \textsc{Batch Prompting}~\citep{cheng2023batch}.
}
\label{fig:inference_comparison}
\vspace{-6mm}
\end{figure}

\section{Introduction}

Large language models (LLMs) capable of following instructions have demonstrated impressive performance across a wide range of tasks~\citep{xu2023wizardlm, openai2023gpt4, anil2023palm, tunstall2023zephyr, wang2023openchat}. However, since LLMs are trained to follow a \textit{single} instruction per inference call, it is questionable whether they also hold the ability to follow complex instructions that necessitate handling \textit{multiple} sub-tasks~\citep{yang2018hotpotqa,geva2021did,cheng2023batch}. Moreover, current evaluation resources are either confined to measuring the LLM's capability in following one-step instructions~\citep{alpaca_eval,vicuna2023,zheng2023judging} or only diagnose the capability to process multi-step instructions in a particular domain such as commonsense reasoning and arithmetic~\citep{geva2021did, cobbe2021training,lightman2023let}.


In this paper, we analyze whether LLMs hold the capability to handle tasks composed of multiple instructions at one inference call, which we denote as \textsc{Multi-Task Inference}. As shown in Figure~\ref{fig:inference_comparison}, we compare the performance and speed with two baselines: (1) \textsc{Single-Task Inference}: addressing sub-tasks sequentially and (2) \textsc{Batch Prompting}: simultaneously processing multiple instances from the same task~\citep{cheng2023batch}. 

For this purpose, we construct the \textsc{MTI Bench} (\textbf{M}ulti-\textbf{T}ask \textbf{I}nference \textbf{Bench}mark), an evaluation dataset featuring 25 tasks, each consisting of 2 to 3 sub-tasks. As shown in Figure~\ref{fig:mti_bench_example}, the \textsc{MTI Bench} is divided into two distinct subsets: (a) the \textsc{Multi-Step} subset, which evaluates the models' ability follow multiple instructions sequentially and (b) the \textsc{Multi-Part} subset, focusing on the models' capability to handle multiple sub-tasks that do not have a sequential dependency. Notably, the \textsc{MTI Bench} sets itself apart from previous multi-hop reasoning~\citep{yang2018hotpotqa, geva2021did} and multi-turn conversation~\citep{zheng2023judging} evaluation suites by providing annotations to assess the \textit{intermediate} performance of LLMs while solving multi-task instructions. This enables researchers to check if LLMs reach the correct answers and evaluate whether their reasoning process is consistent and logical throughout the process.

We evaluate 11 LLMs capable of following instructions, varying in parameter size. Surprisingly, on the \textsc{MTI Bench}, state-of-the-art LLMs such as \textsc{Llama-2-Chat-70B} and \textsc{GPT-4} show up to 7.3\% and 12.4\% better performance with \textsc{Multi-Task Inference} compared to \textsc{Single-Task Inference}.
Moreover, \textsc{Multi-Task Inference} requires x1.46 times less average inference time than \textsc{Single-Task Inference}. These results indicate that users could obtain similar performance with substantially less time when querying instructions that necessitate solving multiple sub-tasks. Through ablation experiments, we suggest that looking at the next sub-task provides critical clues on the answer format for solving the previous sub-task.

Our contributions are as follows:
\begin{itemize}
    \item We are, to the best of our knowledge, the first to develop an evaluation benchmark, the \textsc{MTI Bench}, tailored to analyze the \textsc{Multi-Task Inference} capabilities of LLMs. We fully open-source our code and data.
    \item Our findings demonstrate that \textsc{Multi-Task Inference} surprisingly works well compared to \textsc{Single-Task Inference} only for stronger models. 
    \item We show that \textsc{Multi-Task Inference} offers x1.46 times speed-up compared to \textsc{Single-Task Inference}. This suggests that practitioners can fully leverage the capability of LLMs to solve multiple tasks at one inference call.
\end{itemize}

\section{Related Works}

\subsection{Language Model Evaluation}
While Large Language Models (LLMs) demonstrate impressive performance across a wide range of tasks, it remains essential to assess their properties and behaviors from various perspectives~\citep{chang2023survey,wang2023survey,chia2023instructeval}. Traditionally, evaluations of LLMs primarily focused on performance in specific domains or tasks~\citep{hendrycks2020measuring,srivastava2023beyond}. However, there is a growing interest in holistically evaluating LLMs' properties and high-level capabilities across multiple facets~\citep{liang2022holistic,holtzman2023generative,kim2023prometheus}. Prior research in this area includes measuring overall helpfulness and harmlessness in user interactions~\citep{dubois2023alpacafarm,alpaca_eval,zheng2023judging}, assessing the ability to generate coherent thought chains in reasoning tasks~\citep{fu2023chain, ott2023thoughtsource}, examining the presence of a theory of mind~\citep{zhou2023far, kim2023fantom, mireshghallah2023can}, and evaluating the capacity to avoid producing toxic content~\citep{gehman2020realtoxicityprompts}. In our work, we focus on multi-processing capabilities, specifically the ability of LLMs to process multiple instructions simultaneously, as a novel and significant area to explore and evaluate across various LLMs.

\subsection{Multiprocessing Capabilities of LLMs}

The ability to concurrently process multiple pieces of information is a key indicator of intelligence~\citep{meyer1997computational}. Previous studies have introduced datasets like HotpotQA~\citep{yang2018hotpotqa} and StrategyQA~\citep{geva2021did}, which require multi-hop reasoning. These are designed to train and test the LLM's capability to follow the internal reasoning processes needed for a valid final prediction. However, these datasets do not offer a comprehensive method to assess the accuracy of intermediate steps or to compare concurrent versus sequential processing. Recently, \citet{cheng2023batch} introduced \textsc{Batch Prompting}, aligning with the research direction of our study. However, this approach is limited to examining if LLMs can process multiple instances within the \textit{same} task. In contrast, our \textsc{MTI Bench} encompasses a broader range of scenarios, including instructions comprising multiple sub-tasks that either follow a sequential order (\textsc{Multi-Step} subset) or solve different tasks (\textsc{Multi-Part}).

\section{The \textsc{MTI Bench} Dataset}

In this section, we explain how the \textsc{MTI Bench} is formulated (Section~\ref{sec:3.1}), how we constructed it (Section~\ref{sec:3.2}), and provide an analysis of the diversity, compositionality, and quality. (Section~\ref{sec:3.3}).

\begin{table}[t!]
\centering
\fontsize{7}{8}\selectfont
\begin{tabular}{llccccc}
\toprule
 \multicolumn{1}{c}{} &  & \multicolumn{2}{c}{\# of}            & \multicolumn{1}{c}{} & \multicolumn{2}{c}{Avg. Length} \\ \cmidrule{3-4} \cmidrule{6-7} 
                     &  & Task & \multicolumn{1}{l}{Task Type} &                      & Instruction      & Context      \\ 
\midrule
\textsc{Multi-Step}   &  & 13 & 12 &  & 20.3 & 89.4  \\
\textsc{Multi-Part}   &  & 12 & 16 &  & 22.4 & 104.8 \\
\midrule
\textsc{Total} &  & 25 & 28  &  & 17.4 & 115.8 \\ 
\bottomrule
\end{tabular}%
\caption{\footnotesize Dataset Statistics for \textsc{MTI Bench}. The lengths of instructions and context are measured in the number of words.}
\label{tab:mti_stats}
\end{table}

\begin{figure}[t!]
\includegraphics[width=\columnwidth]{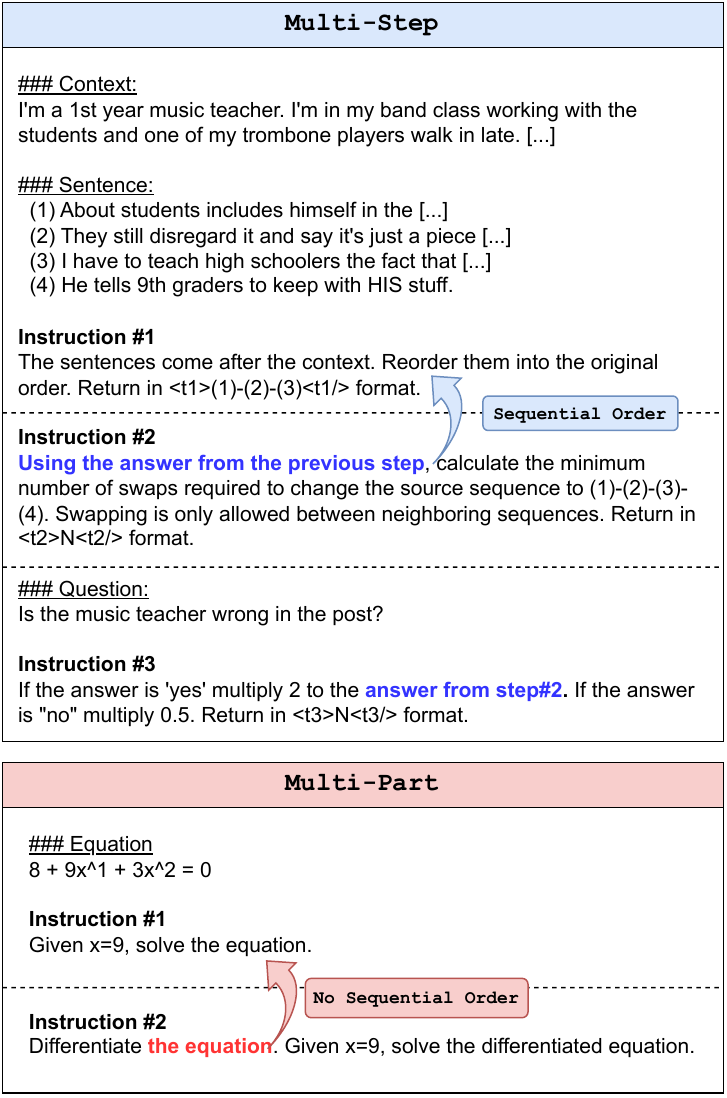}
\centering
\caption{\footnotesize An example comparing the \textsc{Multi-Part} and \textsc{Multi-Step} subset within the \textsc{MTI Bench} dataset. Whereas the \textsc{Multi-Step} necessitates to solve step-by-step since there is a sequential order among the sub-tasks, the sub-tasks within the \textsc{Multi-Part} does not have a sequential order.
}
\label{fig:mti_bench_example}
\end{figure}

\subsection{Task Formulation}\label{sec:3.1}
The \textsc{MTI Bench} (\textbf{M}ulti-\textbf{T}ask \textbf{I}nference \textbf{Bench}mark) is a comprehensive benchmark to evaluate the \textsc{Multi-Task Inference} capabilities of LLMs. The benchmark comprises 25 tasks, each with 200 instances, summing up to 5,000 instances in total. Each task within the benchmark comprises 2 to 3 sub-tasks, selected from a diverse pool of 28 NLP tasks, including Classification, Multiple-Choice Question Answering (MCQA), Arithmetic, and Natural Language Inference. These tasks are divided into two subsets: \textsc{Multi-Step} and \textsc{Multi-Part} containing 13 and 12 tasks respectively.  Five of the 25 tasks consist of 3 sub-tasks. Table~\ref{tab:mti_stats} presents detailed statistics for each subset.

Tasks in the \textsc{Multi-Step} subset demand a sequential approach, with the accuracy of each step being vital for the following ones. This subset assesses LMs' proficiency in managing interdependent tasks. Conversely, the \textsc{Multi-Part} subset consists of contextually related but independent sub-tasks, evaluating LLMs' capacity to process multiple, disparate tasks simultaneously. Both subsets employ exact string matching as the evaluation method, focusing on both intermediate and final accuracy. An example instance for each subset is illustrated in Figure~\ref{fig:mti_bench_example}.

\subsection{Dataset Construction}\label{sec:3.2}

To construct the \textsc{MTI Bench}, we select a wide range of tasks from existing NLP benchmarks. Our primary sources include Quoref~\citep{dasigi2019quoref}, SNLI~\citep{bowman2015large}, MMLU~\citep{hendrycks2020measuring}, and MATH~\citep{hendrycks2021measuring}.  Tables~\ref{ex001}-\ref{ex025} provides a comprehensive list of datasets used to construct the benchmark. The key criteria for source dataset selection are (1) the presence of a rigorous quality control process in the datasets and (2) the potential to integrate the datasets into more complex tasks. The co-authors split into two groups for efficiency: one focused on combining different tasks into composite tasks, while the other screened for and eliminated any combinations that were uninformative or of low quality, subsequently categorizing the tasks into either \textsc{Multi-Step}, or \textsc{Multi-Part} subsets. During the process, 7 out of the initial 32 multi-tasks were deemed unsuitable and removed, resulting in a refined final version of 25 high-quality tasks. Additionally, we crafted a one-shot demonstration for each task, which sequentially resolves the sub-tasks by generating a Chain-of-Thought~\citep{wei2022chain}.

\subsection{Dataset Analysis}\label{sec:3.3}

\begin{table}
\centering
\fontsize{9.5}{11}\selectfont
\begin{tabular}{lccc}
\toprule
Task Type & \multicolumn{1}{l}{\textsc{1st}} & \multicolumn{1}{l}{\textsc{2nd}} & \multicolumn{1}{l}{\textsc{3rd}} \\ 
\midrule
Others & 32\% & 24\% & - \\
Classification & 28\% & 4\% & - \\
Sentence Sorting & 20\% & 12\% & - \\
Answerability Classification & 16\% & 4\% & - \\
Natural Language Inference & 4\% & 8\% & - \\
Extractive QA & - & 4\% & 40\% \\
Arithmetic & - & 16\% & 20\% \\
Multiple-Choice QA & - & 12\% & 20\% \\
Binary QA & - & 4\% & 20\% \\
Wrong Candidate Ranking & - & 8\% & - \\
Judicial Decision & - & 4\% & - \\ \bottomrule
\end{tabular}%
\caption{\footnotesize Distribution of Task Types for each sub-task.}
\label{tab:mti_tasks}
\end{table}

\begin{table}[t!]
\centering
\fontsize{7.5}{9}\selectfont
\begin{tabular}{lccc}
\toprule
Subset & Chi-square Statistic & p-value & Odds Ratio \\ \midrule
\textsc{Multi-Step}         & 394.37  & \( \textbf{$< 0.0001$} \)   & 8.44                \\
\textsc{Multi-Part}        & 128.28  & \( \textbf{$< 0.0001$} \)   & 2.64                \\  \bottomrule
\end{tabular}%
\caption{\footnotesize Chi-squared statistical test results for \textsc{Multi-Step} and \textsc{Multi-Part} subsets. The results indicate that the tasks are properly classified into each category as intended.}
\label{tab:chi}
\end{table}

\paragraph{Diversity} The distribution of NLP tasks in their respective order within the sub-tasks is detailed in Table~\ref{tab:mti_tasks}. No single task type dominates, ensuring a wide-ranging evaluation of model capabilities. There are only five multi-tasks comprised of three sub-tasks, resulting in a relatively constrained diversity for \textsc{3rd} sub-task. 


\paragraph{Compositionality}~To statistically verify the authors' manual classification of multi-tasks into \textsc{Multi-step} and \textsc{Multi-part} subset, we conduct a chi-squared test to study the interdependency within each subset. Initially, a \textsc{GPT-3.5-Turbo} model was used to solve 200 instances of each multi-task combination. Subsequently, a chi-squared test was applied to the outcomes to assess the dependency between the accuracy of each sub-task.  In Table~\ref{tab:chi}, both subsets demonstrated p-values below the 0.01 threshold, refuting the null hypothesis that the sub-tasks are independent. Furthermore, the \textsc{Multi-step} subset features chi-square statistic and odds ratio substantially higher than the \textsc{Multi-part} subset, indicating a more pronounced linear association among its tasks.

\paragraph{Quality} To ensure the quality of the \textsc{MTI Bench}, we conduct a two-step quality check.  Initially, we selected a random sample of eight instructions from each task, making a total of 200 instructions for evaluation. Two of our authors labeled whether each instance showed valid dependencies between sub-tasks and were properly categorized. Tasks were recategorized and rephrased according to the results. After these adjustments, a final round of quality assessment was conducted. This phase involved ten professional annotators, including authors from our team and five externally recruited experts. The hired experts, all master's graduates in finance, business, and computer science, were paid at the rate of \$0.11 per question.

\begin{table}[t!]
\centering
\fontsize{8}{11}\selectfont
\begin{tabular}{ccc}
\toprule
Quality Review Question & \textsc{1st} & \textsc{final} \\ \midrule
\pbox{4.8cm}{\centering \strut  Does the instruction feature valid sub-task dependencies? \strut} & 89\% & 91\% \\
\pbox{4.8cm}{\centering \strut Is the (instruction, context, answer) triplet suitable for the benchmark? \strut} & 88\% & 92\% \\
\pbox{4.8cm}{\centering \strut  Does the task align with its designated category (\textsc{Multi-Step}, \textsc{Multi-Part})?} & 76\% & 88\% \\ 
\midrule
All fields are invalid    &  1\%  &  0\% \\
\bottomrule
\end{tabular}%
\caption{\footnotesize Data quality review for each component within the \textsc{MTI Bench} instance: the instruction, context, answer. Annotators were asked to answer either "Yes" or "No" for each question given a randomly sampled instance from the \text{MTI Bench}. Results show the ratio of "Yes" from the annotators.}
\label{tab:quality}
\end{table}

The evaluation results, presented in Table~\ref{tab:quality}, indicate that after the modification process, majority of the multi-tasks in the benchmark demonstrate valid sub-task dependencies and are correctly categorized. Two annotators reviewed each question, and the Cohen's kappa statistic~\citep{mchugh2012interrater} for inter-annotator agreement on these questions scored 0.82, 0.68, and 0.89, indicating a substantial level of consensus. It was also noted that the remaining misclassifications did not reflect the overall task labeling but were somewhat isolated incidents, likely due to the specific contexts of individual samples. Importantly, even in cases with errors, no instances fail the quality assessment criteria completely, suggesting that the errors were not severe enough to affect the dataset's reliability as a benchmarking tool.

\begin{table*}[t!]
\fontsize{5}{7}\selectfont
\resizebox{\textwidth}{!}{
\begin{tabular}{lccccccccc}
\toprule
\multicolumn{1}{l}{} & \multicolumn{3}{c}{\textsc{Single-Task}}                                       & \multicolumn{3}{c}{\textsc{Batch Prompting}}              & \multicolumn{3}{c}{\textsc{Multi-Task}}                                        \\ \cmidrule(lr){2-4} \cmidrule(lr){5-7} \cmidrule(lr){8-10}
\multicolumn{1}{l}{} & \textsc{M.S.}                   & \textsc{M.P.}                   & \textsc{Average} & \textsc{M.S.}                   & \textsc{M.P.}  & \textsc{Average} & \textsc{M.S.}                   & \textsc{M.P.}  & \textsc{Average} \\
\midrule
\textsc{Tulu-7B} & 0.9 & 0.9 & 0.9 & 0.0 & 0.0 & 0.0 & 0.4 & 1.5 & \underline{\textbf{1.0}} \\
\textsc{Tulu-13B} & 2.9 & 2.6 & 2.8 & 0.0 & 0.0 & 0.0 & 2.1 & 3.8 & \underline{\textbf{3.0}} \\
\textsc{Tulu-30B} & 8.2 & 5.4 & \underline{\textbf{6.8}} & 2.0 & 1.0 & 1.5 & 1.5 & 4.4 & 3.0 \\
\textsc{Tulu-65B} & 1.4 & 4.6 & 3.0 & 2.4 & 3.0 & 2.7 & 5.6 & 7.1 & \underline{\textbf{6.4}} \\
\textsc{Llama-2-Chat-7b} & 2.8 & 4.4 & 3.6 & 1.0 & 0.0 & 0.5 & 5.5 & 7.9 & \underline{\textbf{6.7}} \\
\textsc{Llama-2-Chat-13b} & 1.0 & 3.0 & 2.0 & 0.0 & 0.0 & 0.0 & 2.4 & 4.2 & \underline{\textbf{3.3}} \\
\textsc{Llama-2-Chat-70b} & 8.0 & 9.4 & 8.7 & 7.4 & 8.3 & 7.9 & 16.0 & 20.0 & \underline{\textbf{18.0}} \\
\textsc{Vicuna-7B} & 2.2 & 2.3 & 2.3 & 1.3 & 1.5 & 1.4 & 3.9 & 4.8 & \underline{\textbf{4.4}} \\
\textsc{Vicuna-13B} & 6.5 & 11.6 & \underline{\textbf{9.1}} & 2.4 & 1.9 & 2.2 & 7.3 & 9.3 & 8.3 \\
\midrule
\textsc{GPT-3.5-Turbo} & 18.9 & 23.7 & 21.3 & 18.1 & 19.1 & 18.6 & 21.5 & 26.2 & \underline{\textbf{23.9}} \\
\textsc{GPT-4} & 25.8 & 35.7 & 30.8 & 33.3 & 31.0 & 32.2 & 43.2 & 42.5 & \underline{\textbf{42.9}}\\
\bottomrule
\end{tabular}
}
\caption{\footnotesize Evaluation results of \textsc{Multi-step} (\textsc{M.S.}), and \textsc{multi-part} (\textsc{M.P.}) subset utilizing \textsc{single-task inference}, \textsc{batch-prompting} and \textsc{multi-task inference}. The specified accuracy is the accuracy of correctly completing all sub-tasks (i.e., \textbf{final accuracy}). Evaluations are held in a one-shot setting with chain-of-thought reasoning.  The best comparable
performances among the inference methods are bolded and  underlined.}
\label{tab:main}
\vspace{-3mm}
\end{table*}

\section{Experimental Setup}

In this section, we explain our experimental setup for investigating the \textsc{Multi-Task Inference} capabilities of LLMs. 

\paragraph{Baseline Inference Methods}\label{sec:bim} In addition to \textsc{Multi-Task Inference}, the method in our main consideration, we compare with \textsc{Single-Task Inference} and \textsc{batch prompting}~\citep{cheng2023batch}. Figure~\ref{fig:inference_comparison} illustrates a scenario that compares the three inference methods. Assuming that we are testing an LLM with two instances that consist of 3 sub-tasks, the most naive approach, \textsc{Single-Task Inference} prompts an LLM 6 times, where each inference call corresponds to solve a single sub-task. On the other hand, \textsc{Batch Prompting} groups the same sub-tasks and prompts an LLM to solve multiple instances at once. Lastly, \textsc{Multi-Task Inference} prompts the LLM to solve all the multiple sub-tasks within a single inference call. In general, if $N$ instances consisting of $M$ sub-tasks are given, \textsc{Single-Task Inference} requires $N$ times more inference calls compared to \textsc{Batch Prompting} and $M$ times more inference calls compared to \textsc{Multi-Task Inference}.


\paragraph{Test Models}~We evaluate eleven LLMs capable of following instructions including: (1) \textsc{GPT-4}~\citep{openai2023gpt4}, (2) \textsc{GPT-3.5}~\citep{chatgpt}, (3) \textsc{Tulu} (7b, 13b, 30b, 65b)~\citep{wang2023far}, (4) \textsc{vicuna} (7b, 13b)~\citep{vicuna2023}, and (5) \textsc{Llama-2-Chat} (7b, 13b, 70b)~\citep{touvron2023llama}. For \textsc{GPT-4} and \textsc{GPT-3.5}, we utilize the \textsc{0613} version. Reported results represent the average of three runs, except for \textsc{GPT-4}, which were evaluated in a single run to minimize costs. Open-source models were run using fp16 precision. All evaluations were conducted in a single-shot setting, incorporating Chain-of-Thought reasoning. The hyperparameters used for evaluation are detailed in Appendix~\ref{sec:appendix:inference}.

\paragraph{Evaluation Methodology}~The \textsc{MTI Bench} comprises 28 types of NLP tasks, yielding diverse outputs such as multiple-choice answers, numerical answers(fractional form), and extensive generative responses. Given this variety, directly applying verbalizers like LM-Eval-Harness~\cite{eval-harness} is impractical. Therefore, we prompted LLMs to return their outputs within an HTML tag (e.g., \texttt{<task1>{output}<task1/>}), which is then assessed via exact match (EM).\label{sec:4.3}

\paragraph{Hardware Specifications}~In Section~\ref{sec:5.2}, we examine the inference speed of four models: \textsc{Tulu} (7b, 13b, 30b, 65b)~\citep{wang2023far}. For observation, the hardware configuration for each model size is fixed. Specifically, the \textsc{Tulu} models with \textsc{7b} and \textsc{13b} parameters were tested using a single NVIDIA SXM4 with 80GB RAM. The \textsc{30b} model utilized two of these NVIDIA SXM4 80GB GPU, while the largest, the \textsc{65b} model, was evaluated using eight RTX A6000 with 48GB RAM each. \label{sec:4.4}

\section{Experimental Results}

In this section, we compare \textsc{Single-Task Inference}, \textsc{Batching Prompting} and \textsc{Multi-Task Inference} on the \textsc{MTI Bench} (Section~\ref{sec:5.1}), study the inference latency of each method (Section~\ref{sec:5.2}) and study the efficicacy of \textsc{Multi-Task Inference} on free-form generation (Section~\ref{sec:5.3}).

\subsection{Main Results}\label{sec:5.1}

\begin{figure}[t!]
\includegraphics[width=\columnwidth]{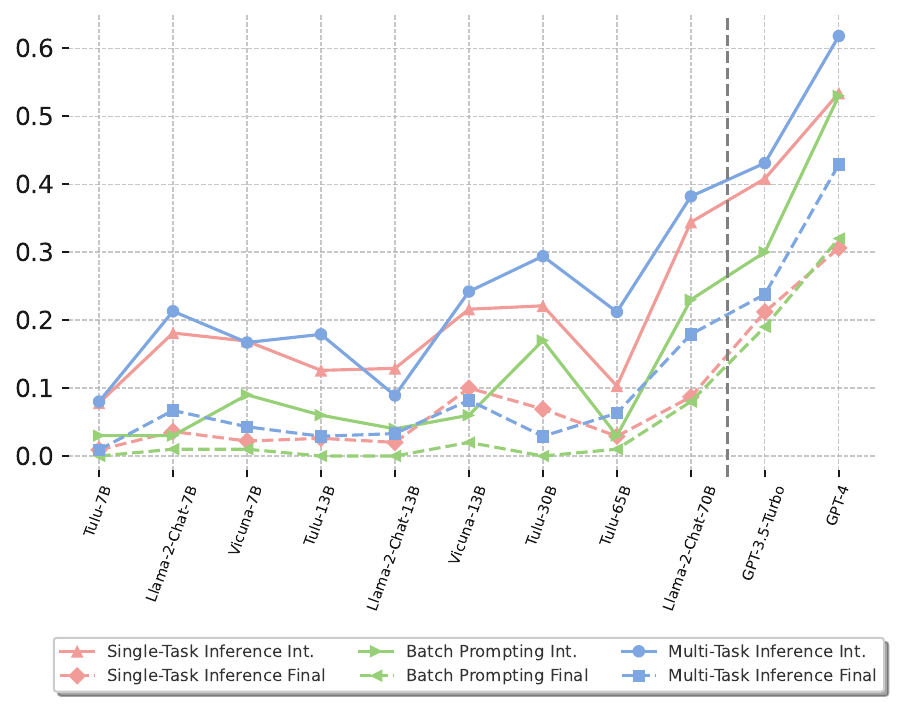}
\centering
\caption{\footnotesize Comparative analysis of LLMs across \textsc{Single-Task Inference} (Green), \textsc{Batch Prompting} (Red), and \textsc{Multi-Task Inference} (Blue). Solid lines represent the models' initial sub-task performance (i.e., \textbf{intermediate accuracy}), while dashed lines indicate their overall accuracy in completing the entire set of tasks (i.e., \textbf{final accuracy}). Models are listed in ascending order by parameter count, with proprietary models listed separately at the end.
}
\label{fig:inference_step}
\vspace{-6mm}
\end{figure}

We first evaluate \textsc{Single-Task Inference}, \textsc{Batching Prompting}, and \textsc{Multi-Task Inference} using the \textsc{MTI Bench}. In Table~\ref{tab:main} we focus on the final accuracy of each model, only considering the cases where it correctly solves the entire combination of sub-tasks. Surprisingly, \textsc{Multi-Task Inference} consistently outperforms the other methods across various models.  Notably, the performance gap between the inference strategies is larger in more powerful models. For instance, with the \textsc{Llama-2-Chat-70B} model, accuracy under \textsc{Single-Task Inference} and \textsc{Batching Prompting} is 8.7\% and 7.9\%, respectively, but it leaps to 16.0\% using \textsc{Multi-Task Inference}. A similar trend is observed in \textsc{GPT-4}, where accuracy escalates from 30.8\% and 32.2\% to 43.2\%. In Figure~\ref{fig:inference_step}, we observe a clear upward scaling trend, which demonstrates that more advanced models exhibit enhanced performance on the \textsc{MTI Bench}, irrespective of the prompting methods employed. This trend suggests that the capability to  concurrently handle multi-task instructions could be an \textit{emergent} property~\citep{wei2022emergent}, associated with the increased scale of models.

The intermediate accuracy for each prompting method is illustrated in Figure~\ref{fig:inference_step}. Notably, \textsc{Multi-Task Inference}, depicted in blue, consistently surpasses alternative prompting methods in both initial and final performances. Furthermore, the efficacy of \textsc{Batch Prompting}, depicted in green, improves as the model size increases, reaching its peak with \textsc{GPT-4}. Despite the improvement, however, a performance gap exists with the remaining inference methods. We conjecture that the performance margin may be tied to the operational nature of \textsc{Batch Prompting}.
It combines multiple tasks without regard to their inter-dependencies, potentially introducing unrelated contexts into a single prompt. This mixing of tasks can confuse the model, as it needs to navigate through irrelevant information multiple times to address the prompt accurately. This observation aligns with existing research that the performance of batching inference improves with model scale \citep{cheng2023batch} and that the presence of non-relevant context can adversely affect model performance~\citep{shi2023large}. 

Finally, the \textsc{MTI Bench} is divided into two subsets: \textsc{Multi-Step} and \textsc{Multi-Part}. As seen in Table~\ref{tab:main}, models generally perform better in the \textsc{Multi-Part} subset. This suggests that inter-task dependency in multi-task instructions is a significant factor that hinders LLM performance, and the ability to manage sequential task dependencies effectively is not uniformly developed across different models.

\begin{table}[t!]
\fontsize{6.7}{9.5}\selectfont{%
\begin{tabular}{lccc} \toprule
Batch Size & \multicolumn{3}{c}{N = 1} \\ \cmidrule(lr){2-4}
Inference Type & \textsc{Single-Task} & \textsc{Batch Prompting} & \textsc{Multi-Task} \\ \midrule
\textsc{Tulu-7B} & 11.3 $\pm$ 5.6 & 5.1 $\pm$ 2.6 & 7.5 $\pm$ 5.2 \\
\textsc{Tulu-13B} & 14.8 $\pm$ 6.0 & 6.3 $\pm$ 2.8 & 9.2 $\pm$ 5.4 \\
\textsc{Tulu-30B} & 51.9 $\pm$ 61.9 & 46.2 $\pm$ 57.3 & 49.2 $\pm$ 42.2 \\
\textsc{Tulu-65B} & 110.1 $\pm$ 54.1 & 52.6 $\pm$ 30.1 & 67.7 $\pm$ 39.6 \\ \bottomrule
\end{tabular}%
}
\caption{\footnotesize The inference latency in solving a multi-task instruction (with a batch size of 1) of the \textsc{Tulu} models measured in seconds. This measurement is an average derived from 1,000 trials.}
\label{isb1}
\end{table}

\begin{table}[t!]
\fontsize{6.7}{9.5}\selectfont{%
\begin{tabular}{lccc} \toprule
Batch Size & \multicolumn{3}{c}{N = 4} \\ \cmidrule(lr){2-4}
Inference Type & \textsc{Single-Task} & \textsc{Batch Prompting} & \textsc{Multi-Task} \\ \midrule
\textsc{Tulu-7B} & 15.4  $\pm$  6.5 & 6.3 $\pm$ 2.8 & 11.3 $\pm$ 6.3 \\
\textsc{Tulu-13B} & 19.4 $\pm$ 6.4 & 7.4 $\pm$ 3.1 & 13.0 $\pm$ 5.6 \\
\textsc{Tulu-30B} & 93.3 $\pm$ 117.7 & 57.6 $\pm$ 65.1 & 63.2 $\pm$ 37.5 \\
\textsc{Tulu-65B} & 156.9 $\pm$ 64.7 & 64.9 $\pm$  33.8 & 96.7 $\pm$ 38.5 \\ \bottomrule
\end{tabular}%
}
\caption{\footnotesize The inference latency in solving a multi-task (with a batch size of 4) of the \textsc{Tulu} models measured in seconds. This measurement is an average derived from 250 trials.}
\label{isb4}
\end{table}

\subsection{Inference Latency}\label{sec:5.2} 
Considering KV caching, intuitively, a model requiring fewer inference calls would be faster in terms of inference speed, assuming it generates an equal number of tokens. Empirically, in Tables~\ref{isb1} and \ref{isb4}, we observe a $1.46 \times$ increase in speed using \textsc{Multi-Task Inference} compared to \textsc{Single-Task Inference}.  This acceleration remains consistent as the batch size increases from 1 to 4. 

Additionally, \textsc{Batch Prompting} demonstrates a $2.1 \times$ increase in speed compared to \textsc{Single-Task Inference}, aligning with the findings in \citep{cheng2023batch}. However, as highlighted in Section~\ref{sec:5.1}, employing \textsc{Batch Prompting} for the \textsc{MTI Bench} results in a marked decrease in performance, making \textsc{Multi-Task Inference} the most viable option.

\subsection{\textsc{Free-Form Generation} Subset}\label{sec:5.3} 

As mentioned in Section~\ref{sec:4.3}, LLMs are prompted to return their outputs within an HTML tag, which are parsed using regular expressions. During our evaluation, we notice that models often struggle to produce outputs in the correct format, potentially skewing their perceived performance. To address this issue, we introduce a new ablation subset called \textsc{Free-Form Generation}. This subset comprises 11 tasks, each divided into two sub-tasks, primarily focused on translation and summarization. Performance evaluation is conducted using the Rouge-L metric. Due to constraints in budget and time, this ablation is narrowed down to assess performance in the following methods: \textsc{Single-Task Inference} and \textsc{Multi-Task Inference}. Further details on the subset are provided in Appendix~\ref{sec:appendix:ffg}.

Table~\ref{table:main_gen} shows the result of our evaluation on the  \textsc{Free-Form Generation} subset. We observe that smaller open-source models tend to perform better with \textsc{Single-Task Inference} outperforming \textsc{Multi-Task Inference}, with margins ranging from 0.02 to 0.15. However, this performance gap narrows for larger open-source models and proprietary models. Notably, for \textsc{GPT-4}, the difference in performance between the two methods is a mere 0.01, indicating that there is no significant difference in their effectiveness regardless of their output formatting.

We conjecture that the slight decrease in the performance of \textsc{Multi-Task Inference} within the \textsc{Free-Form Generation} subset can be attributed to the weaker interdependence of the sub-tasks involved. For example, in task combinations such as translation and summarization, the information provided by the second instruction offers limited insights into solving the first task. This lack of inter-task informational clues may lead to a reduced level of synergy between the tasks, diminishing the benefit of \textsc{Multi-Task Inference} in such scenarios.

\begin{table}[t!]
\centering
\fontsize{7}{9.5}\selectfont{%
\begin{tabular}{lcccc}
\toprule
\multicolumn{1}{c}{\multirow{3}{*}{Models}} & \multicolumn{4}{c}{\textsc{Free-Form Generation}} \\ \cmidrule(lr){2-5}
\multicolumn{1}{c}{} & \multicolumn{2}{c}{\textsc{Single-Task}} & \multicolumn{2}{c}{\textsc{Multi-Task}} \\ \cmidrule(lr){2-3} \cmidrule(lr){4-5}
\multicolumn{1}{c}{} & \textsc{1st} & \textsc{Final} & \textsc{1st} & \textsc{Final} \\
\midrule
\textsc{Tulu-7B} & 0.19 & 0.12 & 0.16 & 0.10 \\
\textsc{Tulu-13B} & 0.17 & 0.21 & 0.15 & 0.12 \\
\textsc{Tulu-30B} & 0.37 & 0.30 & 0.17 & 0.15 \\
\textsc{Tulu-65B} & 0.39 & 0.11 & 0.19 & 0.15 \\
\textsc{Llama-2-Chat-7B} & 0.14 & 0.10 & 0.11 & 0.08 \\
\textsc{Llama-2-Chat-13B} & 0.15 & 0.13 & 0.07 & 0.16 \\
\textsc{Llama-2-Chat-70B} & 0.35 & 0.26 & 0.27 & 0.24 \\
\textsc{Vicuna-7B} & 0.20 & 0.14 & 0.16 & 0.09 \\
\textsc{Vicuna-13B} & 0.22 & 0.18 & 0.23 & 0.16 \\
\midrule
\textsc{GPT-3.5-Turbo} & 0.48 & 0.37 & 0.38 & 0.31 \\
\textsc{GPT-4} & 0.47 & 0.39 & 0.39 & 0.38 \\ 
\bottomrule
\end{tabular}
}
\caption{\footnotesize Evaluation results of \textsc{Free-Form Generation} subset in \textsc{Single-Task Inference} and \textsc{Multi-Task Inference}. Evaluations are held in a one-shot setting. Note that four MCQA tasks are included in this subset as secondary tasks. Performance scores for both the generative and MCQA tasks are calculated using the Rouge-L metric.\protect\footnotemark}
\label{table:main_gen}
\vspace{-3mm}
\end{table}

\footnotetext{\url{https://pypi.org/project/rouge-score/}}

\begin{figure}[t!]
\includegraphics[width=\columnwidth]{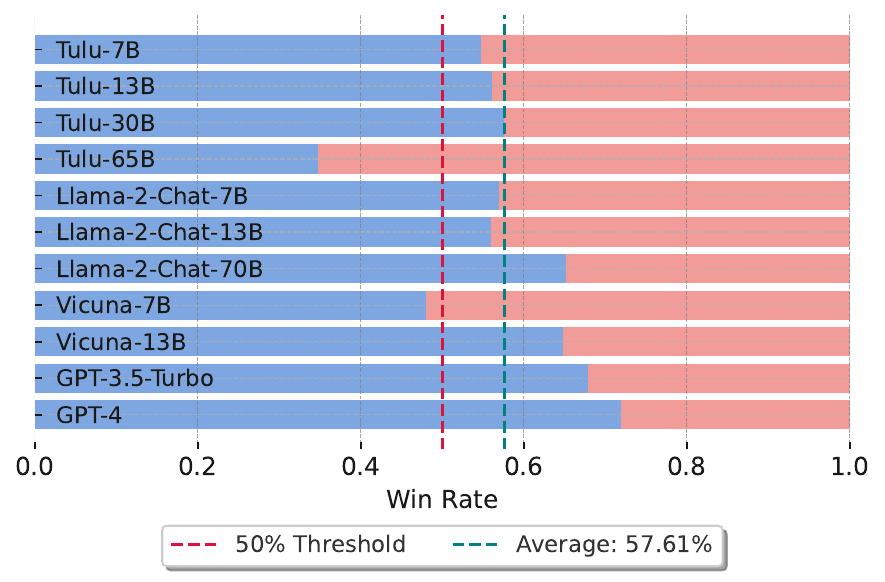}
\centering
\caption{\footnotesize Win Rate Analysis. Blue bars represent \textsc{Multi-Task Inference}  wins, and red bars indicate \textsc{Single-Task Inference}  wins. The green line denotes the average \textsc{Multi-Task Inference} win rate across all models.
}
\label{fig:vote}
\vspace{-6mm}
\end{figure}

In an effort to conduct a more comprehensive comparison between \textsc{Multi-Task Inference} and \textsc{Single-Task Inference} within free-form generation, we conducted further evaluations using the \textsc{MT-Bench}~\citep{zheng2023judging}. A \textsc{GPT-4} model, with the default pairwise comparison prompt from the original paper, was leveraged to judge and select the better response. The results, depicted in Figure~\ref{fig:vote}, reveal that LLMs show a slightly improved performance under \textsc{Multi-Task Inference}, with an average win rate of 58\% across the prompts. Remarkably, \textsc{Llama-2-Chat-70B} and \textsc{GPT-4} under \textsc{Multi-Task Inference} outperformed at 65.2\% and 71.9\% on the prompts, respectively. This shows that the benefit of \textsc{Multi-Task Inference} persists beyond \textsc{MTI Bench} and can be generalized to diverse use cases.


\section{Analysis of \textsc{Multi-Task Inference}}

In our previous section, although it is clear that \textsc{Multi-Task Inference} guarantees speed-up (as explained in Section~\ref{sec:5.2}), it is rather unexpected and surprising that larger models show improved performance on the \textsc{MTI Bench} (Table~\ref{tab:main} and Figure~\ref{fig:inference_step}) and the \textsc{MT Bench} (Figure~\ref{fig:vote}) compared to \textsc{Single-Task Inference}. For a better understanding, we conduct an ablation experiment by inserting additional input components (Section~\ref{sec:6.1}) and conduct a human evaluation to categorize what would be the reason behind the performance improvement (Section~\ref{sec:6.2}).

\begin{table*}[t!]
\centering
\fontsize{10}{14}\selectfont
\begin{tabular}{lcccc}
\toprule
 & \multicolumn{1}{c}{\textsc{Tulu-7B}} & \multicolumn{1}{c}{\textsc{Tulu-13B}} & \multicolumn{1}{c}{\textsc{Tulu-30B}} & \multicolumn{1}{c}{\textsc{GPT-3.5}} \\
 \midrule
\multicolumn{5}{c}{\textsc{Multi-Step}}     \\
\midrule
\textsc{Single-Task Inference}& 8.3 & 15.7 & 35 & 44.6 \\
\textsc{$+$ 2nd Instruction} & 8.5 {\color[HTML]{32CB00} ($+$0.2)} & 25.3  {\color[HTML]{32CB00} ($+$9.6)} & 36.2  {\color[HTML]{32CB00} ($+$1.2) } & 46.3  {\color[HTML]{32CB00} ($+$1.7)} \\
\textsc{$+$ 2nd Context} & 8.7 {\color[HTML]{32CB00} ($+$0.4)} & 20.5 {\color[HTML]{32CB00} ($+$4.8)} & 33.5 {\color[HTML]{CB0000} (-1.5)} & 46.3 {\color[HTML]{32CB00}  ($+$1.7)} \\
\midrule
\multicolumn{5}{c}{\textsc{Multi-Part}}     \\
\midrule
\textsc{Single-Task Inference} & 7.3 & 9.2 & 18.8 & 36.7 \\
\textsc{$+$ 2nd Instruction} & 6.5 {\color[HTML]{CB0000}(-0.8)} & 11.2 {\color[HTML]{32CB00} ($+$2.0)} & 18.0 {\color[HTML]{CB0000}(-0.8)} & 36.0 {\color[HTML]{CB0000}(-0.7)} \\
\textsc{$+$ 2nd Context} & 7.7 {\color[HTML]{32CB00}($+$0.4)} & 11.4 {\color[HTML]{32CB00} ($+$2.2)} & 22.0 {\color[HTML]{32CB00} ($+$3.2)} & 38.7 {\color[HTML]{32CB00} ($+$2.0)} \\
\bottomrule
\end{tabular}%
\caption{\footnotesize Ablation experiment of excluding the \textsc{2nd Instruction} and \textsc{2nd Context}. The specified accuracy represents the models' performance on the first sub-task (denoted as \textbf{intermediate accuracy} in Figure~\ref{fig:inference_step}). Note that \textsc{Single-Task Inference} is excluding both the \textsc{2nd Instruction} and \textsc{2nd Context} compared to \textsc{Multi-Task Inference}.}
\label{tab:ablation_ic}
\end{table*}


\subsection{Ablation Experiment}\label{sec:6.1}
A two-step instance within the \textsc{MTI Bench} would consist of four input components when inferenced via \textsc{Mutli-Task Inference}: (1) 1st Instruction, (2) 1st Context, (3) 2nd Instruction, (4) 2nd Context. On the other hand, when inferenced via \textsc{Single-Task Inference}, only (1) 1st Instruction and (2) 1st Context would be provided as the input during the first inference call. Then on the second inference call, (3) the output of the first inference call, (4) 2nd Instruction, and (5) 2nd Context would be additionally provided.

In Table~\ref{tab:ablation_ic}, we check the effect when adding the 2st Instruction and 2nd Context during \textsc{Single-Task Inference}. Note that \textsc{Single-Task Inference} is the same as excluding both the \textsc{2nd Context} and \textsc{2nd Instruction}. Interestingly, across different models and data subsets, we observe a consistent performance improvement when either the 2nd Instruction or the 2nd Context is provided as additional input when solving the 1st Instruction, indicating a sign of a look-ahead effect.

\subsection{Qualitative Analysis}\label{sec:6.2}

To analyze what kind of look-ahead effect might enable language models to show improved performance on the first instruction, we conduct a qualitative analysis by checking the 107 instances where \textsc{GPT-4} correctly solves using \textsc{Multi-Task Inference} but not with \textsc{Single-Task Inference}. Interestingly, we discover the following four patterns that supplement the look-aheading behavior of LMs: (1) \textbf{No Outputs}: \textsc{Single-Task Inference} provided no output, suggesting there were no viable answers. Conversely, \textsc{Multi-Task Inference}, while acknowledging the implausibility of all answers, still opts to select one. (2) \textbf{Multiple Outputs}: \textsc{single-task Inference} offered multiple answers, whereas the \textsc{Multi-Task Inference} approach selected the most relevant one. (3) \textbf{Referencing}: \textsc{Multi-Task Inference} leveraged information from a subsequent task to enhance its response to the initial task. (4) \textbf{Planning}: \textsc{Multi-Task Inference} appeared to plan its solution before addressing the task.

\begin{table}
\centering
\fontsize{8.5}{11}\selectfont
\begin{tabular}{lc}
\toprule
                 & \multicolumn{1}{l}{Observed Instances \%} \\ 
\midrule
No Outputs & 25\%                                      \\
Multiple Outputs       & 8\%                                       \\
Referencing        & 6\%                                       \\
Planning         & 3\%     \\
\bottomrule
\end{tabular}%
\caption{\footnotesize Qualitative assessment results of \textsc{GPT-4} outputs; The remaining 58\% show no specific patterns.}
\label{tab:gpt4_quality}
\end{table}

Patterns 1 and 2 highlight the role of \textsc{Multi-Task Inference} in providing a form of external feedback. The existence of subsequent tasks indicates whether an answer exists, thereby eliciting a response from the model. Conversely, Patterns 3 and 4 demonstrate that \textsc{Multi-Task Inference} enables LLMs to utilize their full context window. This broader context usage, which extends beyond the immediate task, allows for more comprehensive problem-solving. The frequency of each pattern from our qualitative assessment is provided in Table~\ref{tab:gpt4_quality}. Sample instances of the observed patterns are provided as Figure~\ref{fig:qual}.

\section{Conclusion}

In this work, we present the \textsc{MTI Bench}, a  comprehensive benchmark consisting of 5,000 instances spanning 25 diverse tasks, designed to assess the capability of LLMs in simultaneous multi-tasking. Our analysis within the benchmark compares \textsc{Multi-Task Inference}, \textsc{Single-Task Inference} and \textsc{Batch Prompting}. The results indicate a superior performance by \textsc{Multi-Task Inference}, despite reduced inference steps and a 1.46-fold increase in speed, demonstrating its efficiency in handling concurrent tasks.

\section{Limitations}
In this work, we try our best to offer a broad range of analyses, yet there are limitations that future studies should consider. 

First, the \textsc{MTI Bench} predominantly focuses on English, with the \textsc{Free Form Generation} ablation subset, adding French, and German. This linguistic range falls short of encompassing the wide diversity of different dialects and languages. 

Second, the source dataset for \textsc{MTI Bench} is largely oriented towards academic benchmarks. This focus might restrict its applicability in more general, user-oriented contexts. Future iterations should consider integrating more varied datasets to better mirror the multifaceted nature of everyday language use. 

Third, another significant area concerns the automatic evaluation of model performance. Although our work employs a variety of methods such as model-based evaluation, exact matching, and Rouge-L, there is a need for additional studies on alignment with human preferences. 

Fourth, the \textsc{MTI Bench} only has a test set since the motivation was to test the \textsc{Multi-Task Inference} capabilities of language models. Yet, it would be an interesting direction to see if the smaller models that underperformed in this work could improve their multi-processing capabilities by training on data instances with a similar format as the instances in the \textsc{MTI Bench}. 

Lastly, we only conducted our experiments in a one-shot setting. This was primarily because we observed that smaller models exhibit near zero accuracy when tested in a zero-shot setting and including more than two demonstrations resulted in too long input length. Yet, we acknowledge the importance of examining the impact of including additional demonstrations since models that support longer input lengths are gradually being introduced. We view this as a promising future research direction.
\bibliography{anthology,custom}
\clearpage
\appendix

\section{MTI Bench}
\label{sec:appendix:mti}

Tables \ref{ex001}-\ref{ex025} provide a comprehensive overview of the 25 multi-tasks featured in the \textsc{MTI Bench}. Each table includes the category, sub-tasks, and the original dataset of each multi-task. Furthermore, an example is provided to help a better understanding of the benchmark. Please note that for some examples, the context has been abbreviated for better readability.

\section{Inference Details}
\label{sec:appendix:inference}

During our experiments, we use the hyperparameters as shown in Table~\ref{tab:config_eval}.

\begin{table}[ht]
\fontsize{8}{10}\selectfont
\centering
\begin{tabular}{cccc}
\toprule
Temperature & Top-p & Repetition Penalty   & Max Output Length\\
\midrule
0.7 & 1.0 & 1.0 & 2048\\
\bottomrule
\end{tabular}
\caption{Hyperparameters used for experiments in \textsc{MTI Bench}.}
\label{tab:config_eval}
\end{table}

\section{\textsc{Free Form Generation} Subset}
\label{sec:appendix:ffg}

Apart from \textsc{multi-step} and \textsc{multi-part} subsets discussed in Section~\ref{sec:3.1}, we introduce a  \textsc{free-form generation} subset in our ablation studies detailed in Section~\ref{sec:5.3}. This subset follows the same creation process as the original MTI Bench, except for the hired annotators for quality assessment. 

\begin{table}[ht]
\resizebox{\columnwidth}{!}{%
\begin{tabular}{ll}
\toprule
\# of Tasks & 12 \\  
\# of Instances & 2,400 \\ \midrule
Language & EN, FR, DE \\ 
List of Sub-Tasks & Translation, Summarization, MCQA \\ \midrule
Source Dataset & \parbox{0.8\columnwidth}{FLORES-200~\cite{nllb2022} \\ Belebele~\citep{bandarkar2023belebele} \\ Wikilingua~\citep{ladhak-etal-2020-wikilingua}} \\ 
\bottomrule
\end{tabular}%
}
\caption{Details for the \textsc{Free Form Generation} Subset}
\end{table}

\section{Examples for Section~\ref{sec:6.2}}

\begin{figure*}[t!]
\includegraphics[width=0.9\textwidth]{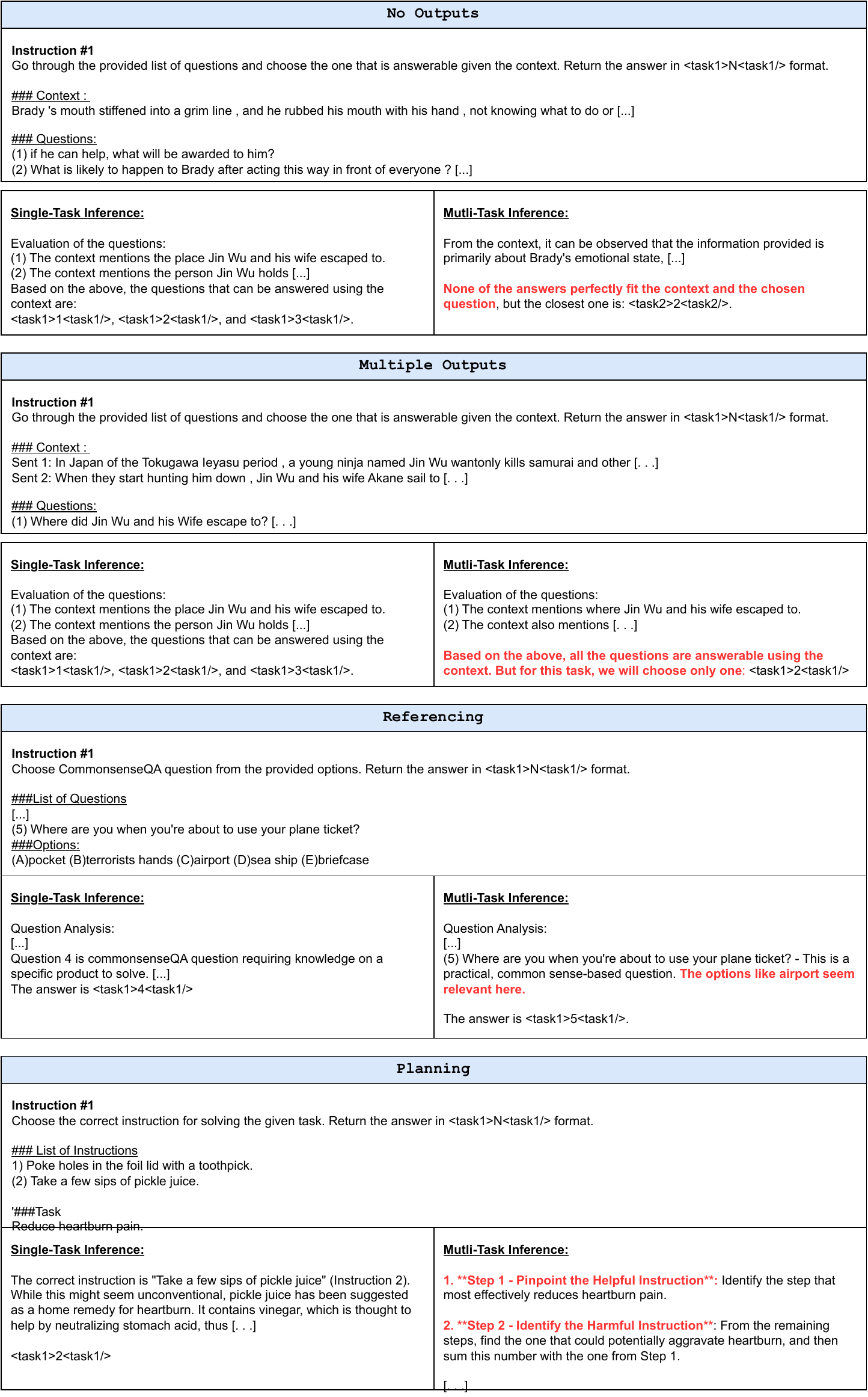}
\centering
\caption{\footnotesize Examples of \textbf{No Outputs}, \textbf{Multiple Outputs},\textbf{Referencing} and \textbf{Planning} used by \textsc{GPT-4} during the \textsc{multi-task inference}.
}
\label{fig:qual}
\end{figure*}

In this section we provide sample instances for the following patterns discussed at Section~\ref{sec:6.2}: No Outputs, Multiple Outputs, Referencing, and Planning. See Figure~\ref{fig:qual}.

\begin{table*}[ht]
\fontsize{7}{8}\selectfont
\resizebox{\textwidth}{!}{%
\begin{tabular}{ll}
\toprule
Task ID & 001 \\ \midrule
Category & \textsc{Multi-Step} \\ \midrule
Sub-Tasks & Answerability Classification - Extractive Question Answering \\ \midrule
Source Dataset & \textsc{Quoref}~\citep{dasigi2019quoref} \\ \midrule
Example & \parbox{0.85\textwidth}{Read the following passage, and follow the given steps,\\ \#1 Go through the provided list of questions and choose the one that is answerable given the context. Return the answer in \textless{}task1\textgreater{}N\textless{}task1/\textgreater format.\\ \#2 Answer the question you have chosen in step \#1. Return the answer in \textless{}task2\textgreater{}N\textless{}task2/\textgreater format.\\ \\ \#\#\# Context: Passage: Big Butte Creek drains approximately 245 square miles (635 km2) of southern Oregon. {[}. . .{]}\\ \#\#\# List of Questions: \\  (1) What watershed is split into two geographic regions?\\ (2) What two entities was the foundation split into in october 2016?\\ (3) What century was Europe split into two city states and kingdoms?\\ (4) How many years was Nashua split into two cities?\\ (5) Who likes to divide their projects into relevent time periods and geographic regions?} \\ 
\bottomrule
\end{tabular}%
}
\caption{Multi-Task 001 from the \textsc{MTI Bench}.}
\label{ex001}
\end{table*}

\begin{table*}[ht]
\fontsize{7}{8}\selectfont
\resizebox{\textwidth}{!}{%
\begin{tabular}{ll}
\toprule
Task ID & 002 \\ \midrule
Category & \textsc{Multi-Step} \\ \midrule
Sub-Tasks & Sentence Sorting - Bubble Sorting - Binary Question Answering \\ \midrule
Source Dataset & \textsc{SCRUPLES}~\citep{lourie2021scruples} \\ \midrule
Example & \parbox{0.85\textwidth}{Read the following passage, and follow the given steps. \\ \#1 The list of sentences come after the context. Reorder them to its original order. Return the answer in \textless{}task1\textgreater{}(1)-(2)-(3)-(4)\textless{}task1/\textgreater format.\\ \#2 Use your answer for step\#1 as a source sequence. Calculate the minimum number of swaps required to change the source sequence to (1)-(2)-(3)-(4). Swapping is only allowed between neighboring sequences. Return the answer in \textless{}task2\textgreater{}N\textless{}task2/\textgreater format.\\ \#3 Read the reordered text and solve the question. If the answer is "yes" multiply 2 to the answer for step\#2. If the answer is "no" multiply 0.5. Return the answer in \textless{}task3\textgreater{}N\textless{}task3/\textgreater format.\\ \\ \#\#\# Context: **TL;DR My mother died this week, my girlfriend started a fight days later over my inability to talk about it, then she {[}. . .{]}\\ \#\#\# List of sentences:\\ (1) Or else am I justified in feeling this way?\\ (2) I do try very hard to be self-critical, so if you all think I'm in the wrong here, need to just let this go or even apologize to her to prevent things from getting worse, then please tell me.\\ (3) To me, she's being unnecessarily heated and petty on an awful week for me.\\ (4) If she decides not to come, I feel like that might be the end of our relationship.\\ \#\#\# Question: Is my girlfriend wrong in the post ?} \\ 
\bottomrule
\end{tabular}%
}
\caption{Multi-Task 002 from the \textsc{MTI Bench}.}
\label{ex002}
\end{table*}

\begin{table*}[ht]
\fontsize{7}{8}\selectfont
\resizebox{\textwidth}{!}{%
\begin{tabular}{ll}
\toprule
Task ID & 003 \\ \midrule
Category & \textsc{Multi-Step} \\ \midrule
Sub-Tasks & Answerability Classification - Multiple-Choice Question Answering \\ \midrule
Source Dataset & \textsc{COSMOS QA}~\citep{huang2019cosmos} \\ \midrule
Example & \parbox{0.85\textwidth}{Read the following passage, and follow the given steps,\\ \#1 Go through the provided list of questions and choose the one that is answerable given the context. Return the answer in \textless{}task1\textgreater{}N\textless{}task1/\textgreater format.\\ \#2 Choose the correct answer for the question you have chosen in step \#1. Return the answer in \textless{}task2\textgreater{}N\textless{}task2/\textgreater format.\\     \\ \#\#\# Context: Two cats ( one is an itty bitty kitty that they bought a couple of days ago ) , {[}. . .{]}\\ \#\#\# List of Questions: \\ (1) What makes it possible for humans to live on other planets?\\ (2) What may be the reason there are so many people living in a small apartment?\\ {[}. . .{]}  \\ \#\#\# Answer Choice: \\ (1) The individual makes pictures all the time .\\ (2) We all split the bills and makes it easier to live .\\ {[}. . .{]}}\\ 
\bottomrule
\end{tabular}%
}
\caption{Multi-Task 003 from the \textsc{MTI Bench}.}
\label{ex003}
\end{table*}

\begin{table*}[ht]
\fontsize{7}{8}\selectfont
\resizebox{\textwidth}{!}{%
\begin{tabular}{ll}
\toprule
Task ID & 004 \\ \midrule
Category & \textsc{Multi-Part} \\ \midrule
Sub-Tasks & Answerability Classification - Answer \& Question Matching \\ \midrule
Source Dataset & \textsc{DROP}~\citep{dua2019drop} \\ \midrule
Example & \parbox{0.85\textwidth}{Read the following passage, and follow the given steps,\\ \#1 Go through the provided list of questions and choose all that is answerable given the context. Return the answer in \textless{}task1\textgreater{}{[}N, N, ..{]}\textless{}task1/\textgreater format.\\ \#2 From the questions selected at task\#1 choose the one that best suits the given answer. Return the answer in \textless{}task2\textgreater{}N\textless{}task2/\textgreater format.\\ \\ \#\#\# Context: Passage: Until 1998, Shearer was paid \$30,000 per episode. During a pay dispute in 1998, [. . .]\\ \#\#\# List of Questions: \\ (1) Which year was the 400,000 salary per episode cut down by 100,000?\\ (2) How many more dollars did voice actors receive in 2008 than they negotiated for in 2004?\\ (3) How many years after taking the throne for himself and refusing to pay tribute did a military response begin?\\ (4)  How many years after receiving a raise did Shearer take a pay cut?\\ (5) How many students does \$16,000 a year pay for?\\ \#\#\# Answer: 3.
}\\ 
\bottomrule
\end{tabular}%
}
\caption{Multi-Task 004 from the \textsc{MTI Bench}.}
\label{ex004}
\end{table*}

\begin{table*}[ht]
\fontsize{7}{8}\selectfont
\resizebox{\textwidth}{!}{%
\begin{tabular}{ll}
\toprule
Task ID & 005 \\ \midrule
Category & \textsc{Multi-Part} \\ \midrule
Sub-Tasks & Question \& Context Matching - Wrong Candidate Ranking \\ \midrule
Source Dataset & \textsc{cosmos qa}~\citep{huang2019cosmos} \\ \midrule
Example & \parbox{0.85\textwidth}{Read the following passage, and follow the given steps,\\ \#1 Read the following list of text and determine which one contains the answer to the question. Return the answer in \textless{}task1\textgreater{}N\textless{}task1/\textgreater format.\\ \#2 Read the list of wrong candidates provided determine which one serves as the best wrong answer for the question. Return the answer in \textless{}task2\textgreater{}N\textless{}task2/\textgreater format.\\ \\ \#\#\# Question: What does the narrator think about the video game they were playing ? \\ \#\#\# List of Text: \\ (1) The walk in was quite tiring actually plus the hot scorching sun. {[}. . .{]}\\  (2) So basically the lecture was on when to know if the guy is a nutcase or not. {[}. . .{]}\\  (3) I almost cried when I saw the mud in the arena , it was fucking insane! {[}. . .{]}\\ \\ \#\#\# Wrong Candidates: \\ (1) She wants a PC.\\ (2) Because it stopped running Firefox .\\ (3) They lost it at school .\\ (4) It could be a lot better .\\ (5) They were taking a fitness test at the gym .}\\ 
\bottomrule
\end{tabular}%
}
\caption{Multi-Task 005 from the \textsc{MTI Bench}.}
\label{ex005}
\end{table*}

\begin{table*}[ht]
\fontsize{7}{8}\selectfont
\resizebox{\textwidth}{!}{%
\begin{tabular}{ll}
\toprule
Task ID & 006 \\ \midrule
Category & \textsc{Multi-step} \\ \midrule
Sub-Tasks & Answerability Classification - Necessary Sentence Identification \\ \midrule
Source Dataset & \textsc{multi rc}~\citep{khashabi-etal-2018-looking} \\ \midrule
Example & \parbox{0.85\textwidth}{Read the following passage, and follow the given steps,\\ \#1 Go through the provided list of questions and choose the one that is answerable given the context. Return the answer in \textless{}task1\textgreater{}N\textless{}task1/\textgreater format.\\ \#2 Choose sentences from the context that is necessary to answer the question you have chosen in step \#1. Return the answer in \textless{}task2\textgreater{}{[}N, N, ..{]}\textless{}task2/\textgreater format.\\ \\ \#\#\# Context: \\  Sent 1: The film opens with Sunita , a medical student , and her friends working on a project about the human brain.\\ Sent 2: She wants to investigate the curious case of Sanjay Singhania , a notable city businessman , who is reported to have anterograde amnesia.\\ Sent 3: Her professor denies access to Sanjay 's records as it is currently under criminal investigation. {[}. . .{]}\\ \\ \#\#\# List of Questions: \\ (1) can a person function with half a brain\\ (2) Sunita is working on a project about the human brain and wants to interview which person with anterograde amnesia?\\ (3) Beyonce did an interview with which magazine and was asked about feminism?\\ (4) What is anterograde amnesia?\\ (5) Why is the writer working on a project?}\\ 
\bottomrule
\end{tabular}%
}
\caption{Multi-Task 006 from the \textsc{MTI Bench}.}
\label{ex006}
\end{table*}

\begin{table*}[ht]
\fontsize{7}{8}\selectfont
\resizebox{\textwidth}{!}{%
\begin{tabular}{ll}
\toprule
Task ID & 007 \\ \midrule
Category & \textsc{Multi-Part} \\ \midrule
Sub-Tasks & Sentence Sorting - Inappropriate Question Identification \\ \midrule
Source Dataset & \textsc{multi rc}~\citep{khashabi-etal-2018-looking} \\ \midrule
Example & \parbox{0.85\textwidth}{Read the following passage, and follow the given steps,\\ \#1 The provided list of sentences come after the provided context, order the properly. Return the answer in \textless{}task1\textgreater{}(1)-(2)-(3)-(4)\textless{}task1/\textgreater format.\\ \#2 Choose one question that cannot be answered with the context. Return the answer in \textless{}task2\textgreater{}N\textless{}task2/\textgreater format.\\ \\ \#\#\# Context: \\ Preservation and Conservation: In 1857 the Great Western Railway Company built a main line to Scotland, {[}. . .{]}\\ \#\#\# List of Sentences: \\ 1: In 1974 a total reorganization of local government throughout the UK did away with the old counties of Cumberland and Westmoreland and created the larger county of Cumbria.\\ 2: While the Lake District encourages and welcomes visitors, its popularity can damage the landscape and tax local transportation services. {[}. . .{]}\\ \\ \#\#\# List of Questions: \\ 1: What 1879 event caused a group of concerned individuals to form the Lake District Defense Association?\\ 2: What organization was a precursor to the National Trust? [. . .]}\\ 
\bottomrule
\end{tabular}%
}
\caption{Multi-Task 007 from the \textsc{MTI Bench}.}
\label{ex007}
\end{table*}

\begin{table*}[ht]
\fontsize{7}{8}\selectfont
\resizebox{\textwidth}{!}{%
\begin{tabular}{ll}
\toprule
Task ID & 008 \\ \midrule
Category & \textsc{Multi-Part} \\ \midrule
Sub-Tasks & Sentence Sorting - Answer \& Question Matching \\ \midrule
Source Dataset & \textsc{ropes}~\citep{lin2019reasoning} \\ \midrule
Example & \parbox{0.85\textwidth}{Read the following passage, and follow the given steps,\\ \#1 The provided list of sentences come after the provided context, order them properly. Return the answer in \textless{}task1\textgreater{}(1)-(2)-(3)-(4)\textless{}task1/\textgreater format.\\ \#2 Choose one question that best suits the given passage and answer. Return the answer in \textless{}task2\textgreater{}N\textless{}task2/\textgreater format.\\ \\ \#\#\# Context: \\ New species develop naturally through the process of natural selection. {[}. . .{]}\\ \#\#\# List of Sentences: \\ (a): Mike lives in a cold mid-western city, where there is not much predator prey interaction.\\ (b): He also knew that darker coats are more suitable in cold environment with less predator prey interaction. {[}. . .{]}\\ \\ \#\#\# List of Questions: \\ 1. Which squirrels would most likely reproduce in greater numbers, lighter or darker?\\ 2. Would the color be darker or lighter at point B than at point A? {[}. . .{]}\\ \\ \#\#\# Answer: greater.}\\ 
\bottomrule
\end{tabular}%
}
\caption{Multi-Task 008 from the \textsc{MTI Bench}.}
\label{ex008}
\end{table*}

\begin{table*}[ht]
\fontsize{7}{8}\selectfont
\resizebox{\textwidth}{!}{%
\begin{tabular}{ll}
\toprule
Task ID & 009 \\ \midrule
Category & \textsc{Multi-Step} \\ \midrule
Sub-Tasks & Necessary Sentence Identification - Sentence Sorting \\ \midrule
Source Dataset & \textsc{TIMETRAVEL}~\citep{qin2019counterfactual} \\ \midrule
Example & \parbox{0.85\textwidth}{Read the following passage, and follow the given steps,\\ \#1 Choose one sentence that does not originally belong to the passage. Return the answer in \textless{}task1\textgreater{}N\textless{}task1/\textgreater format.\\ \#2 Reorganize the remaining sentences into its original order. Return the answer in \textless{}task2\textgreater{}(1)-(2)-(3)-(4)\textless{}task2/\textgreater format.\\ \\ \#\#\# List of Sentences: \\ (1) My daughter jumped up and grabbed the blue one out of her hand\\ (2) Nana chased her down, caught her, and tickled her until she laughed\\ (3) She took off running down the hall while waving the sock in the air\\ (4) She held up an orange sock and a blue one.\\ (5) Nana came into the room with a puzzled look on her face.\\ (6) She held up an orange shirt and a blue one.}\\
\bottomrule
\end{tabular}%
}
\caption{Multi-Task 009 from the \textsc{MTI Bench}.}
\label{ex009}
\end{table*}

\begin{table*}[ht]
\fontsize{7}{8}\selectfont
\resizebox{\textwidth}{!}{%
\begin{tabular}{ll}
\toprule
Task ID & 010 \\ \midrule
Category & \textsc{Multi-Step} \\ \midrule
Sub-Tasks & Coherent Passage Detection - Sentence Sorting \\ \midrule
Source Dataset & \textsc{abductivenli}~\citep{bhagavatula2019abductive} \\ \midrule
Example & \parbox{0.85\textwidth}{Read the following passage, and follow the given steps,\\ \#1 You will be given five group of sentences. Only one of them is a group of coherent sentences. The others include an injected sentence. Find the coherent passage. Return the answer in \textless{}task1\textgreater{}N\textless{}task1/\textgreater format.\\ \#2 Reorganize the passage you chose in step 1 into its original order. Return the answer in \textless{}task2\textgreater{}(1)-(2)-(3)-(4)\textless{}task2/\textgreater format.\\ \\ \#\#\# List of Sentences: \\ 1.\\ (1) Jackson now lives with the guilt of being a thief.\\ (2) Mark kept the wallet.\\ (3) Jackson stole a wallet at a party on Friday.\\ 2.\\ (1) The teacher also gave the lab partner detention for not doing anything.\\ (2) The lab partner sat there like they knew everything.\\ (3) The instructor announced the lab that we're going to perform.[. . .] } \\
\bottomrule
\end{tabular}%
}
\caption{Multi-Task 010 from the \textsc{MTI Bench}.}
\label{ex010}
\end{table*}

\begin{table*}[ht]
\fontsize{7}{8}\selectfont
\resizebox{\textwidth}{!}{%
\begin{tabular}{ll}
\toprule
Task ID & 011 \\ \midrule
Category & \textsc{Multi-Step} \\ \midrule
Sub-Tasks & Question Classification - Multiple Choice Question Answering \\ \midrule
Source Dataset & \textsc{commonsenseqa}~\citep{talmor-etal-2019-commonsenseqa} \\ \midrule
Example & \parbox{0.85\textwidth}{Read the following questions, and follow the given steps,\\ \#1 Choose one question that best suits a "CommonsenseQA" dataset. Return the answer in \textless{}task1\textgreater{}N\textless{}task1/\textgreater format.\\ \#2 Read the options and solve the question you chose at step\#1. Return the answer in \textless{}task1\textgreater{}N\textless{}task1/\textgreater format.\\ \\ \#\#\# List of Questions: \\ (1) What does the client think about the house? \\ (2) Where would you put uncooked crab meat?\\ (3) Why did the man buy dog food at the supermarket?\\ (4) \_, 52, earned about \$94million in salary during his 16 seasons in the National Basketball Association.\\ (5) Question: What is Hector Hammond's job?} \\
\bottomrule
\end{tabular}%
}
\caption{Multi-Task 011 from the \textsc{MTI Bench}.}
\label{ex011}
\end{table*}

\begin{table*}[ht]
\fontsize{7}{8}\selectfont
\resizebox{\textwidth}{!}{%
\begin{tabular}{ll}
\toprule
Task ID & 012 \\ \midrule
Category & \textsc{Multi-Part} \\ \midrule
Sub-Tasks & Sentence Sorting - Answerability Classification - Extractive Question Answering\\ \midrule
Source Dataset & \textsc{squad 1.1}~\citep{rajpurkar2016squad} \\ \midrule
Example & \parbox{0.85\textwidth}{Read the following text, and follow the given steps,\\ \#1 Reorder the given sentences to its original order. Return the answer in \textless{}task1\textgreater{}(1)-(2)-(3)-(4)\textless{}task1/\textgreater format.\\ \#2 Go through the provided list of questions and choose the one that is answerable given the context. Return the answer in \textless{}task2\textgreater{}N\textless{}task2/\textgreater format.\\ \#3 Solve the question you have chose from step\#2. Extract the answer from the passage of step\#1. Return the answer in \textless{}task3\textgreater{}N\textless{}task3/\textgreater format.\\ \\ \#\#\# List of Sentences: \\ (1) The flowers tended to grow in a spiral pattern, to be bisexual (in plants, this means both male and female parts on the same flower), and to be dominated by the ovary (female part).\\ (2) The most primitive flowers probably had a variable number of flower parts, often separate from (but in contact with) each other.\\ {[}. . .{]}\\ \\ \#\#\# List of Questions: \\ (1) Who'd tactic evolved?\\ (2) When do they plant yams and millet?\\ (3) What did some plant parts do as they evolved?\\ (4) what became more mammal-like as they evolved?\\ (5) What did some plant parts do when the flower had only male parts?} \\
\bottomrule
\end{tabular}%
}
\caption{Multi-Task 012 from the \textsc{MTI Bench}.}
\label{ex012}
\end{table*}

\begin{table*}[ht]
\fontsize{7}{8}\selectfont
\resizebox{\textwidth}{!}{%
\begin{tabular}{ll}
\toprule
Task ID & 013 \\ \midrule
Category & \textsc{Multi-Part} \\ \midrule
Sub-Tasks & Answer \& Question Matching -  Wrong Candidate Ranking\\ \midrule
Source Dataset & \textsc{piqa}~\citep{bisk2020piqa} \\ \midrule
Example & \parbox{0.85\textwidth}{Read the following text, and follow the given steps,\\ \#1 Choose the correct answer for the given question. Return the answer in \textless{}task1\textgreater{}N\textless{}task1/\textgreater format.\\ \#2 Choose the best incorrect answer for the given question. Return the answer in \textless{}task2\textgreater{}N\textless{}task2/\textgreater format.\\ \\ \#\#\# List of Answers: \\ (1) Using a fork, stir the pecan mixture with the butter until evenly coated. Press pecan butter mixture into the bottom of your springform pan.\\ (2) If the semolina mixture is too dry, you can add a few teaspoons of milk until it reaches the right consistency\\ (3) Using a pie plate, stir the pecan mixture with the butter until evenly coated. Press pecan butter mixture into the bottom of your springform pan.\\ (4) Heat up milk in the colander until it is 105 degrees, then add yeast and a pinch of sugar to the bowl of milk\\ (5) Take some boiled milk in a small bowl and add the saffron strands to it and watch the saffron turn the milk yellow.\\ \\ \#\#\# Question: \\ How do I add the pecan mixture in the pan when making creamy chocolate toffee torte? } \\
\bottomrule
\end{tabular}%
}
\caption{Multi-Task 013 from the \textsc{MTI Bench}.}
\label{ex013}
\end{table*}

\begin{table*}[ht]
\fontsize{7}{8}\selectfont
\resizebox{\textwidth}{!}{%
\begin{tabular}{ll}
\toprule
Task ID & 014 \\ \midrule
Category & \textsc{Multi-Step} \\ \midrule
Sub-Tasks & Classification  -  Arithmetic\\ \midrule
Source Dataset & \textsc{COM2SENSE}~\citep{singh2021com2sense} \\ \midrule
Example & \parbox{0.85\textwidth}{Read the following text, and follow the given steps,\\ \#1 Read through the following list of sentences and choose all sentences that are plausible and matches commonsense. Return the answer in \textless{}task1\textgreater{}{[}N, N,...{]}\textless{}task1/\textgreater format.\\ \#2 Count the number of inplausible sentences and express its ratio in fraction form. Return the answer in \textless{}task2\textgreater{}n/N\textless{}task2/\textgreater format.\\ \\ \#\#\# List of Sentences: \\ (1) Natalie was embarrassed when her husband yelled at her in the store, so she told all her classmates about the experience.\\ (2) It is better to have white wine with fish than red wine\\ (3) Ricki was delighted to see that 2 customers came to her opening night. [. . .] } \\
\bottomrule
\end{tabular}%
}
\caption{Multi-Task 014 from the \textsc{MTI Bench}.}
\label{ex014}
\end{table*}

\begin{table*}[ht]
\fontsize{7}{8}\selectfont
\resizebox{\textwidth}{!}{%
\begin{tabular}{ll}
\toprule
Task ID & 015 \\ \midrule
Category & \textsc{Multi-Step} \\ \midrule
Sub-Tasks & Classification -  Arithmetic  -  Arithmetic\\ \midrule
Source Dataset & \textsc{winowhy}~\citep{zhang2020winowhy} \\ \midrule
Example & \parbox{0.85\textwidth}{Read the following text, and follow the given steps,\\ \#1 Read through the following list of sentences and choose all sentences that are incorrect reasons for the given question. Return the answer in \textless{}task1\textgreater{}{[}N, N, ..{]}\textless{}task1/\textgreater format. \\ \#2 Count the number correct reasons and express its ratio in fraction form. Return the answer in \textless{}task2\textgreater{}n/N\textless{}task2/\textgreater format.\\ \#3 Solve the following equation: (ratio\_of\_correct\_reason) add (ratio\_of\_wrong\_reason) Write in decimal form. Return the answer in \textless{}task3\textgreater{}N\textless{}task3/\textgreater format.\\ \\ \#\#\# Question:\\ Sentence: Carol believed that Rebecca suspected that she had stolen the watch. Question: Why does the 'she' refer to carol?\\ \#\#\# List of Sentences: \\ (1) Because If Rebecca regrets something of course she must of been the one that stole the watch.\\ (2) Because Because rebecca wouldn't suspect herself in a crime, she would know.\\ (3) Because Rebecca was known to have been in an abusive relationship with Carol. [. . .]} \\
\bottomrule
\end{tabular}%
}
\caption{Multi-Task 015 from the \textsc{MTI Bench}.}
\label{ex015}
\end{table*}

\begin{table*}[ht]
\fontsize{7}{8}\selectfont
\resizebox{\textwidth}{!}{%
\begin{tabular}{ll}
\toprule
Task ID & 016 \\ \midrule
Category & \textsc{Multi-part} \\ \midrule
Sub-Tasks & Classification -  Classification  -  Multiple Choice Question Answering\\ \midrule
Source Dataset & \textsc{Argument Facet Similarity Corpus}~\citep{misra-etal-2016-measuring} \\ \midrule
Example & \parbox{0.85\textwidth}{Read the following text, and follow the given steps,\\ \#1 Read through the following list of texts. The topic of each text is one of the following: (1) death\_penalty (2) gun\_control (3) gay\_marriage. Choose all text that suits the death\_penalty topic. Return the answer in \textless{}task1\textgreater{}{[}N, N, ..{]}\textless{}task1/\textgreater format.\\ \#2 The type of each text is one of the following: (1) argument\_similarity (2) argument\_clarity. Out of the text you have chose in step\#1 choose argument\_clarity text. Return the answer in \textless{}task2\textgreater{}N\textless{}task2/\textgreater format.\\ \#3 Solve the question you have chose in step\#2. Choose from: (1) Similar (2) Not Similar (3) Valid (4) Ivalid. Return the answer in \textless{}task3\textgreater{}N\textless{}task3/\textgreater format.\\ \\ List of Texts:\\ (1) Sent1: Since heterosexuals are provided the means to have a happy marriage and homosexuals are not, homosexuals are not equal to heterosexuals.\\  Sent2: Allowing straight marriage to provide for U.S. citizenship, while gays have no option (marriage or civil union).\\ (2) Well, if that's a reason to ban homosexuals from marriage, then along the same line of thought, then any couple that is infertile or chooses not to have children should not be permitted to get married.\\ (3) Sent1: The judge may or may not feel the death penaly is warranted.\\  Sent2: Many people find some crimes heinous enough to warrent the death penalty.[. . .]} \\
\bottomrule
\end{tabular}%
}
\caption{Multi-Task 016 from the \textsc{MTI Bench}.}
\label{ex016}
\end{table*}

\begin{table*}[ht]
\fontsize{7}{8}\selectfont
\resizebox{\textwidth}{!}{%
\begin{tabular}{ll}
\toprule
Task ID & 017 \\ \midrule
Category & \textsc{Multi-part} \\ \midrule
Sub-Tasks & Sentence Sorting - Binary Question Answering\\ \midrule
Source Dataset & \textsc{mcscript}~\citep{ostermann-etal-2018-mcscript} \\ \midrule
Example & \parbox{0.85\textwidth}{Read the following text, and follow the given steps,\\ \#1 The list of sentences come after the context. Reorder them to its original order. Return the answer in \textless{}task1\textgreater{}(1)-(2)-(3)-(4)\textless{}task1/\textgreater format.\\ \#2 Choose the best answer for the given question. Return the answer in \textless{}task2\textgreater{}N\textless{}task2/\textgreater format.\\ \\ \#\#\# Context:\\ I find that cats are very good about reminding you when it is time for them to eat. They will meow and often stand by their bowl. {[}. . .{]}\\ List of Sentences:\\ (1) So the first thing I do is head to the kitchen to see if there is an open can of her food in the refrigerator.\\ (2) I am careful to measure her food so that she gets just a quarter cup of wet and a quarter cup of dry because I don't want her to be overweight.\\ (3) Next I 'll go to my pantry and pull out a bag of her favorite dry food and mix a little of each into her food bowl.\\ (4) Then I 'll take the time to make sure she has plenty of water before I set her dish on the floor for her to begin eating.\\ \\ \#\#\# Question:\\ What is taken from the kitchen cupboard?\\ Options: 1: measuring cup  2: Bag of cat food.} \\
\bottomrule
\end{tabular}%
}
\caption{Multi-Task 017 from the \textsc{MTI Bench}.}
\label{ex017}
\end{table*}

\begin{table*}[ht]
\fontsize{7}{8}\selectfont
\resizebox{\textwidth}{!}{%
\begin{tabular}{ll}
\toprule
Task ID & 018 \\ \midrule
Category & \textsc{Multi-step} \\ \midrule
Sub-Tasks & Necessary Sentence Identification - Sentence Sorting - Extractive Question Answering\\ \midrule
Source Dataset & \textsc{duorc}~\citep{DuoRC} \\ \midrule
Example & \parbox{0.85\textwidth}{Read the following passage, and follow the given steps.\\ \#1: The list of sentences come after the context. Choose one that does not original belong to the context. Return the answer in \textless{}task1\textgreater{}N\textless{}task1/\textgreater format.\\ \#2: Reorder the remaining into its original order. Return the answer in \textless{}task2\textgreater{}(1)-(2)-(3)-(4)\textless{}task2/\textgreater format.\\ \#3: Answer the given question. Return the answer in \textless{}task3\textgreater{}N\textless{}task3/\textgreater format.\\ \\ \#\#\# Context\\ Deepak (Shashi Kapoor) is on trial for the murder of his wealthy wife Vimla, but is acquitted and set free. {[}. . .{]}\\ \\ \#\#\# List of Sentences\\ (1) Soon, Sapna learns that Gopal's real name is Deepak, who was previously accused of murdering his first wife.\\ (2) Gopal's ever changing behavior throws everyone into suspicion and Sapna fears she will be his next victim.\\ (3) Is Gopal innocent or Guilty?\\ {[}. . .{]}\\ \#\#\# Question\\ Who believes that Gopal is annoying and is stalking her?} \\
\bottomrule
\end{tabular}%
}
\caption{Multi-Task 018 from the \textsc{MTI Bench}.}
\label{ex018}
\end{table*}

\begin{table*}[ht]
\fontsize{7}{8}\selectfont
\resizebox{\textwidth}{!}{%
\begin{tabular}{ll}
\toprule
Task ID & 019 \\ \midrule
Category & \textsc{Multi-part} \\ \midrule
Sub-Tasks & Natural Language Inference - Natural Language Inference\\ \midrule
Source Dataset & \textsc{SNLI}~\citep{bowman-etal-2015-large} \\ \midrule
Example & \parbox{0.85\textwidth}{Read the following text, and follow the given steps,\\ \#1 Determine the relationship. between sentences 1\&2. Choose from: (1) Entailment (2) Contradiction (3) Neutral. Return the answer in \textless{}task1\textgreater{}N\textless{}task1/\textgreater format.\\ \#2 Choose between the given list of sentences that replaces sentence 2 and make a entailment relationship with sentence 1. Return the answer in \textless{}task2\textgreater{}N\textless{}task2/\textgreater format.\\ \\ \#\#\# Sentence 1: An older man, dressed in red, yellow, and black, is standing outside waving a large flag and a long horn. \\ \#\#\# Sentence 2: An older man is standing outside waving to a car driving past.\\ \\ \#\#\# List of Sentences:\\ (A) An older man is proudly waving a large American flag.\\ (B) There is a man outdoors waving a flag.} \\
\bottomrule
\end{tabular}%
}
\caption{Multi-Task 019 from the \textsc{MTI Bench}.}
\label{ex019}
\end{table*}

\begin{table*}[ht]
\fontsize{7}{8}\selectfont
\resizebox{\textwidth}{!}{%
\begin{tabular}{ll}
\toprule
Task ID & 020 \\ \midrule
Category & \textsc{Multi-part} \\ \midrule
Sub-Tasks & Classification - Natural Language Inference\\ \midrule
Source Dataset & \textsc{MNLI}~\citep{williams-etal-2018-broad} \\ \midrule
Example & \parbox{0.85\textwidth}{Read the following text, and follow the given steps,\\ \#1 Classify the given statements to one of the following categories : 1. FACE-TO-FACE, 2. GOVERNMENT, 3. LETTERS, 4. 9/11, 5. SLATE, 6. TELEPHONE, 7. TRAVEL, 8. VERBATIM, 9. OUP, 10. FICTION. Choose all that fits in category 5. Return the answer in \textless{}task1\textgreater{}{[}N, N, .. {]}\textless{}task1/\textgreater format.\\ \#2 Choose a sentence that is in an entailment relationship with the statement you chose in step\#1. If their are two or more answer for step\#1 use the first one. Return the answer in \textless{}task2\textgreater{}N\textless{}task2/\textgreater format.\\ \\ List of Statements:\\ (1) yes but yes and i kind of have always pooh-poohed military educations but i think that for this kid {[}. . .{]}\\ (2) He was pro-German, as he would have been pro-Boer.\\ (3) Historian Thomas Reeves believes that, despite the media's reluctance to look into Kennedy's private life, if he had lived to have a second {[}. . .{]}\\ \\ List of Sentences:\\  1. This kid is not very well behaved or smart.\\ 2. I generally don't like the idea of military educations.\\ 3. I fully support military educations for kids.} \\
\bottomrule
\end{tabular}%
}
\caption{Multi-Task 020 from the \textsc{MTI Bench}.}
\label{ex020}
\end{table*}

\begin{table*}[ht]
\fontsize{7}{8}\selectfont
\resizebox{\textwidth}{!}{%
\begin{tabular}{ll}
\toprule
Task ID & 021 \\ \midrule
Category & \textsc{Multi-part} \\ \midrule
Sub-Tasks & Algebra - Differentiation\\ \midrule
Source Dataset & \textsc{Super Natural Instructions - task 090}~\citep{wang2022super} \\ \midrule
Example & \parbox{0.85\textwidth}{Read the following passage, and follow the given steps. \\ \#1 Solve the given equation: \(3+8x^1+6x^2\), x=10. Return the answer in \textless{}task1\textgreater{}N\textless{}task1/\textgreater format.\\ \#2 Differentiate the equation from step\#1 Solve the equation. Return the answer in \textless{}task2\textgreater{}N\textless{}task2/\textgreater format.} \\
\bottomrule
\end{tabular}%
}
\caption{Multi-Task 021 from the \textsc{MTI Bench}.}
\label{ex021}
\end{table*}

\begin{table*}[ht]
\fontsize{7}{8}\selectfont
\resizebox{\textwidth}{!}{%
\begin{tabular}{ll}
\toprule
Task ID & 022 \\ \midrule
Category & \textsc{Multi-Step} \\ \midrule
Sub-Tasks & Prime Classification - Arithmetic\\ \midrule
Source Dataset & \textsc{Super Natural Instructions - task 092}~\citep{wang2022super} \\ \midrule
Example & \parbox{0.85\textwidth}{Read the following passage, and follow the given steps. \\
\#1 Choose all prime numbers: (1) 99028 (2) 41549 (3) 51481 (4) 94135. Return the answer in <task1>[N, N, ...]<task1/> format. \\
\#2 Sum your choices at step\#1.Return the answer in <task2>N<task2/> format.} \\
\bottomrule
\end{tabular}%
}
\caption{Multi-Task 022 from the \textsc{MTI Bench}.}
\label{ex022}
\end{table*}

\begin{table*}[ht]
\fontsize{7}{8}\selectfont
\resizebox{\textwidth}{!}{%
\begin{tabular}{ll}
\toprule
Task ID & 023 \\ \midrule
Category & \textsc{Multi-Step} \\ \midrule
Sub-Tasks & Classification - Arithmetic\\ \midrule
Source Dataset & \textsc{math}~\citep{hendrycks2021measuring} \\ \midrule
Example & \parbox{0.85\textwidth}{Read the following passage, and follow the given steps. \\ \#1 Read through the given questions. Each question fall into one of the following categories. Choose a question of measurement category. Return the answer in \textless{}task1\textgreater{}N\textless{}task1/\textgreater format.\\ \#2 Solve the question you have chose at step\#1. Return the answer in \textless{}task2\textgreater{}N\textless{}task2/\textgreater format.\\ \\ \#\#\# List of Questions:\\ (1) What is prob of picking 1 h and 2 p when three letters picked without replacement from \{h: 1, e: 3, p: 2, n: 6, q: 1\}?\\ (2) Let p = 182843/22 $+$ -8316. Calculate the common denominator of 70/32 - (1 $+$ -1) and p.\\ (3) How many milliseconds are there in 38.5396 microseconds?\\ (4) Let y(a) = -a $+$ 5. Let m be y(3). Solve f $+$ 16 = -0*f - 4*c, -3*c - 12 = -m*f for f.\\ (5) Calculate (3/(-6))/(33/(-44)).} \\
\bottomrule
\end{tabular}%
}
\caption{Multi-Task 023 from the \textsc{MTI Bench}.}
\label{ex023}
\end{table*}

\begin{table*}[ht]
\fontsize{7}{8}\selectfont
\resizebox{\textwidth}{!}{%
\begin{tabular}{ll}
\toprule
Task ID & 024 \\ \midrule
Category & \textsc{Multi-Step} \\ \midrule
Sub-Tasks & Classification - Multiple Choice Question Answering\\ \midrule
Source Dataset & \textsc{mmlu}~\citep{hendrycks2020measuring} \\ \midrule
Example & \parbox{0.85\textwidth}{Read the following passage, and follow the given steps. \\ \#1 Read through the given questions. Choose one question that is high school level. Return the answer in \textless{}task1\textgreater{}N\textless{}task1/\textgreater format.\\ \#2 Solve the question you have chose at step\#1. Return the answer in \textless{}task2\textgreater{}N\textless{}task2/\textgreater format.\\ \\ List of Questions:\\ (1) A discrete graph is complete if there is an edge connecting any pair of vertices. How many edges does a complete graph with 10 vertices have?\\ (A)10 (B)20 (C)25 (D)45\\ (2) When n = 11, what is the value of 10 – (n $+$ 6)?\\ (A)–7 (B)5 (C)7 (D)27\\ (3) Find the area of the first quadrant region bounded by y = x\textasciicircum{}2, y = cos(x), and the y-axis.\\ (A)0.292 (B)0.508 (C)0.547 (D)0.667} \\
\bottomrule
\end{tabular}%
}
\caption{Multi-Task 024 from the \textsc{MTI Bench}.}
\label{ex024}
\end{table*}

\begin{table*}[ht]
\fontsize{7}{8}\selectfont
\resizebox{\textwidth}{!}{%
\begin{tabular}{ll}
\toprule
Task ID & 025 \\ \midrule
Category & \textsc{Multi-part} \\ \midrule
Sub-Tasks & Mask Infilling - Judicial Decision\\ \midrule
Source Dataset & \textsc{casehold}~\citep{zheng2021does} \\ \midrule
Example & \parbox{0.85\textwidth}{Read the following passage, and follow the given steps. \\ \\ \#1 Read through the given text. Choose one phrase that best suits the blank({[}MASK{]}) in the text. Return the answer in \textless{}task1\textgreater{}N\textless{}task1/\textgreater format.\\ \#2 Given the text choose the correct judicial decision. Return the answer in \textless{}task2\textgreater{}N\textless{}task2/\textgreater format.\\ \\ \#\#\# Context:\\ Denial of his motions in Leo, Mr. Leo filed his second suit in the Kansas federal district court (No. 10-2495-JTM), {[}. . .{]} In a memorandum and order dated March 22, 2011, 2011 WL {[}MASK{]} {[}MASK{]} {[}MASK{]} {[}MASK{]} concluded that {[}. .  .{]}\\ \\ \#\#\# List of Phrase:\\ (1) contemporaneous objections. See, e.g.,\\ (2) the reality and the\\ (3) substantially certain to result\\ (4) 1097759, the district court\\ (5) the City bears the\\ \\ \#\#\# List of Judicial Decisions\\ (1) holding that although pro se briefs are liberally construed even pro se litigants must brief arguments in order to preserve them\\ (2) holding that pro se pleadings are to be liberally construed [. . .]} \\
\bottomrule
\end{tabular}%
}
\caption{Multi-Task 025 from the \textsc{MTI Bench}.}
\label{ex025}
\end{table*}

\end{document}